\pdfoutput=1

\documentclass[11pt]{article}

\usepackage[final]{acl}

\usepackage{times}
\usepackage{latexsym}

\usepackage[T1]{fontenc}

\usepackage[utf8]{inputenc}

\usepackage{microtype}

\usepackage{inconsolata}

\usepackage{graphicx}

\usepackage{amsmath,amssymb,amsfonts}

\usepackage{booktabs}
\usepackage{subcaption}
\usepackage{xspace}
\usepackage{arydshln}
\usepackage[most]{tcolorbox}

\usepackage{multirow}
\usepackage{threeparttable}

\newcommand{\ourname}{{LMTransplant}\xspace} 

\begin{document}


\title{Transplant Then Regenerate: A New Paradigm for Text Data Augmentation}

\author{Guangzhan Wang$^1$, Hongyu Zhang$^2$, Beijun Shen$^1$, Xiaodong Gu$^1$\textsuperscript{\footnotemark[1]} \\
$^1$Shanghai Jiao Tong University \quad $^2$Chongqing University\\
\texttt{\{wangguangzhan, bjshen, xiaodong.gu\}@sjtu.edu.cn } \\
}
\author{
Guangzhan Wang$^{\spadesuit}$ \,
Hongyu Zhang$^{\heartsuit}$ \,
Beijun Shen$^{\spadesuit}$ \,
Xiaodong Gu$^{\spadesuit}$\thanks{Correspondence: \texttt{xiaodong.gu@sjtu.edu.cn}} \\[4pt]
$^{\spadesuit}$\,\textit{Shanghai Jiao Tong University}
\quad
$^{\heartsuit}$\,\textit{Chongqing University}\\[4pt]
\texttt{\{wangguangzhan, bjshen, xiaodong.gu\}@sjtu.edu.cn}\\
\texttt{hyzhang@cqu.edu.cn}
}

\maketitle
\thispagestyle{plain}

\begin{abstract}
Data augmentation is a critical technique in deep learning. Traditional methods like Back-translation typically focus on lexical-level rephrasing, which primarily produces variations with the same semantics. 
While large language models (LLMs) have enhanced text augmentation by their “knowledge emergence” capability, controlling the style and structure of these outputs remains challenging and requires meticulous prompt engineering. 
In this paper, we propose \textit{\ourname}, a novel text augmentation paradigm leveraging LLMs. The core idea of \ourname is \textit{transplant-then-regenerate}: incorporating seed text into a context expanded by LLM, and asking the LLM to regenerate a variant based on the expanded context. 
This strategy allows the model to create more diverse and creative content-level variants by fully leveraging the knowledge embedded in LLMs, while preserving the core attributes of the original text. 
We evaluate \ourname across various text-related tasks, demonstrating its superior performance over existing text augmentation methods. Moreover, \ourname demonstrates exceptional scalability as the size of augmented data grows.

\end{abstract}

\section{Introduction}
Data augmentation is a critical technique in deep learning \cite{khosla2020enhancing, DA_DL_2021, DA_DL_2022}. Deep learning models are data-hungry and often suffer from limited datasets. Data augmentation generates additional training samples through transforming or rephrasing the existing dataset. 
This process increases data diversity, reduces the risk of overfitting, and enhances the models' generalization ability.

Data augmentation has been extensively studied in NLP tasks \cite{ feng-etal-2021-survey, 2023-survey, chen-etal-2023-survey, DA-2023-survey}. One simple prior approach \cite{wei-zou-2019-eda} enhances text data through word-level transformations, such as random insertion and deletion. While easy to implement, it often generates low-quality samples that severely disrupt the semantic coherence of the generated text. 
Later, Back-translation \cite{2015Improving} has been commonly used for sentence level rephrasing. Specifically, a translation model first translates the original text into a different language and then translates it back into the original language. Although the augmented text is semantically coherent, it often exhibits high similarity to the original text, leading to poor data diversity \cite{2023-survey, understanding}.

\begin{figure}[!t]
    \centering
    \includegraphics[width=3in, trim=0 0 0 0 clip]{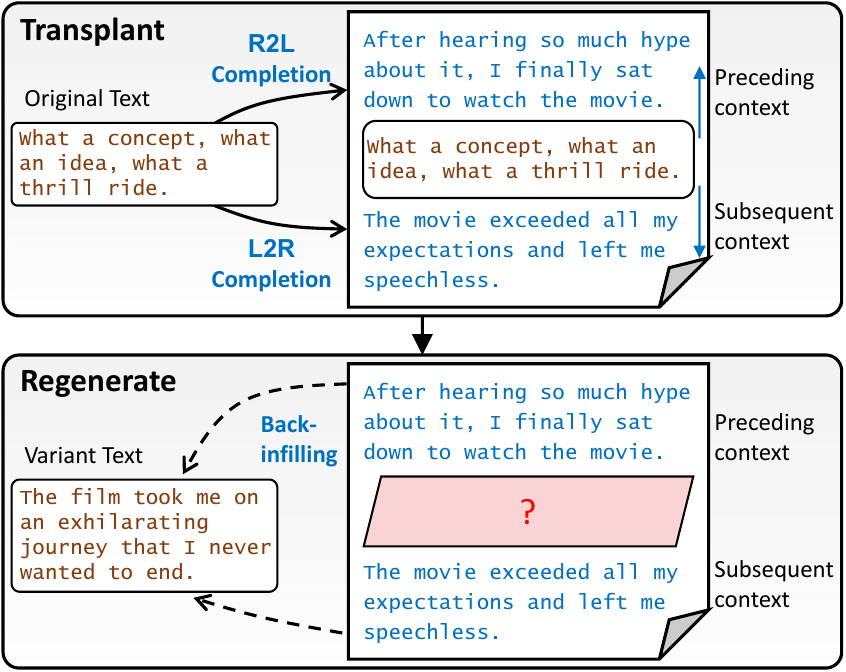}
    \caption{Illustration of \ourname.}
    \label{fig_overview}
\end{figure}

Recent advancements in LLMs have spurred significant interest in LLM-based data augmentation methods \cite{whitehouse-etal-2023-llm, ding-etal-2024-data, zhou2024survey, qiao2024autoact}. 
A key characteristic of LLMs is their ``knowledge emergence'' capability, which stems from two factors: (1) the extensive prior knowledge acquired during pre-training, and (2) their robust language understanding and instruction-following abilities \cite{CoDa, ABEX}. 
These strengths allow LLMs to generate desired outputs directly through demonstrations or natural language instructions without requiring additional training.
A notable example is AugGPT \cite{dai2023auggpt}, which instructs ChatGPT to rephrase text for improving text classification performance. However, this rephrasing-based method only generates variants with similar semantics and underutilizes the rich knowledge embedded in LLMs, limiting the diversity and creativity of the generated data. As a result, when augmented data volume reaches a certain threshold, further increasing may not yield performance improvements \cite{zhou2024survey}.
Although contextual augmentation methods can enhance content diversity, such as GPT3Mix \cite{yoo-etal-2021-gpt3mix-leveraging}, which leverages the powerful few-shot learning capabilities of LLMs to generate mixed augmented text, it is highly sensitive to example quality and input order, potentially introducing uncontrolled biases that are not aligned with the original data distribution.

To address these limitations and better leverage the potential of LLMs, we propose \ourname, a novel text data augmentation paradigm. The core idea of \ourname is \textit{transplant-then-regenerate} (TTR), namely, embedding the original text into an expanded context generated by LLMs and then instructing LLMs to regenerate a variant based on this enriched context.
Specifically, \ourname employs bidirectional text continuation---backward (right-to-left) and forward (left-to-right)---to create the preceding and subsequent context of the original text.
This original text is then masked within its expanded context, and LLMs are prompted to generate a replacement that 
introduces novel content diversity beyond rephrasing, while still preserving core attributes of the original text. 
Therefore, \ourname enhances both the diversity and creativity of the generated text, while maintaining alignment with the original data distribution.

We apply \ourname to various deep learning tasks, including text classification, question answering and named entity recognition (NER), and compare its performance with existing data augmentation methods. Experimental results demonstrate that \ourname can generate higher-quality augmented data. Training models with data augmented by our approach yields significant performance improvements across all tasks. Compared to non-augmentation, \ourname achieves accuracy gains of 28.16\%, 19.96\%, 7.68\%, 23.66\%, and 10.25\% on the SST-2, TREC, SNIPS, MLQA, and CoNLL-2003 datasets, respectively. 



In summary, our contributions are as follows:

\begin{itemize}
\setlength{\itemsep}{0pt}
\setlength{\topsep}{0pt}
\setlength{\parsep}{0pt}
    \item  We propose a novel transplant-based paradigm for text data augmentation. Unlike existing methods which primarily focus on rephrasing, \ourname crafts content-level text variants, thereby crafting higher-quality augmented texts.
    \item  We present a novel transplant and regeneration algorithm using bidirectional text continuation and masked text prediction. 
    The algorithm allows for generating 
    core attributes similar yet more diverse and creative text 
    by effectively utilizing knowledge embedded in LLMs.
    Experiments demonstrate that, \ourname achieves significant performance improvements across different tasks.

\end{itemize}

\section{Approach}
\label{sec:approach}

\subsection{Overview}

We propose \ourname, a novel data augmentation paradigm. 
The core idea of \ourname is \textit{transplant-then-regenerate}: integrating the original text into contextual scenarios generated by LLMs, and then asking LLMs to regenerate new variants given expanded contexts. This strategy allows the model to create content-level variants while preserving the core attributes of the original text.
Figure \ref{fig_overview} illustrates the entire augmentation process. For a given text, \ourname uses bidirectional text continuation to generate its preceding and subsequent contexts (Section~\ref{ss:BidirectionalCompletion}). 
Subsequently, \ourname masks the original text in the transplanted text and asks the LLM to regenerate the missing parts given the crafted contexts, thereby producing new variants of the original text (Section~\ref{ss:IntermediateContentPrediction}). 
Each step of this process will be elaborated in the following sections.

\subsection{Transplant}
\label{ss:BidirectionalCompletion}



Given a seed text, we incorporate it into a relevant contextual scenario. Specifically, we treat the seed text as a fragment of a broader contextual passage, and then use an LLM to generate a semantically natural and logically coherent surrounding context. This process can be conceptualized as bidirectional text continuation, which involves two steps: 
(1) a forward (left-to-right) continuation process that continues writing the subsequent context of the seed text, and 
(2) a backward (right-to-left) continuation process that reconstructs the preceding context of the seed text.

\begin{figure}[!t]
    \centering
    \includegraphics[width=3in]{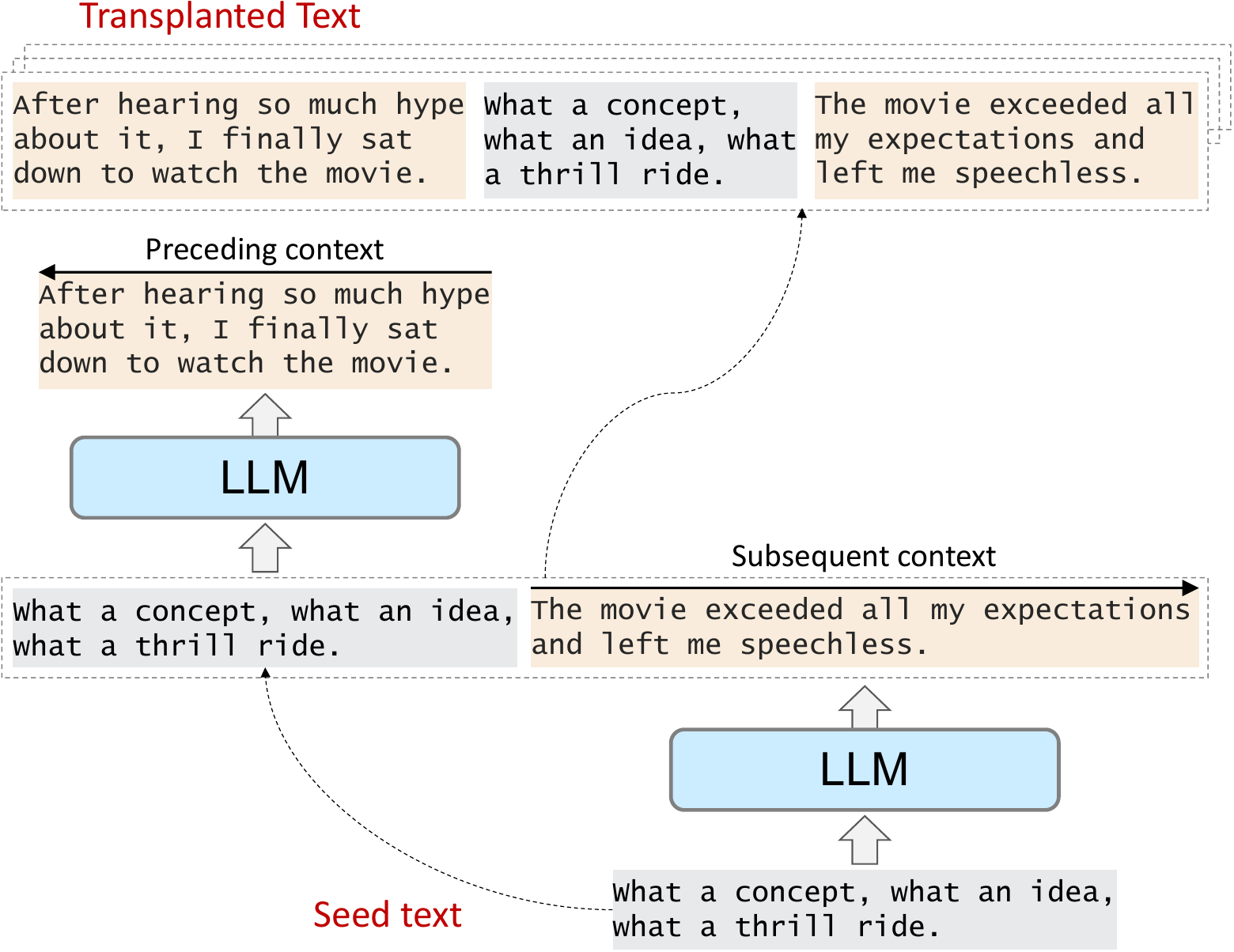}
    \caption{Illustration of text transplant.}
    \label{bidirectional_completion}
\end{figure}

Figure \ref{bidirectional_completion} illustrates the bidirectional text continuation process. 
First, the LLM generates a subsequent text that naturally extends the seed text, combining both to form an expanded passage.
Next, the LLM generates a preceding text based on this expanded input. 
This process ensures that the generated context remains thematically aligned with the original text, while introducing novel information other than rephrasing through the knowledge embedded in LLMs, such as new words and expressions.
Leveraging the language understanding and instruction-following capabilities of LLMs, we employ prompt engineering to guide content generation. The prompt template used for transplanting is provided in Appendix \ref{sec:PromptTemplates}.

\subsection{Regeneration}
\label{ss:IntermediateContentPrediction}

The transplant phase generates multiple contextual scenes bearing the original text. 
We then introduce a \textit{regeneration} process, where we prompt the LLM to regenerate new text variants that seamlessly integrate into the expanded context. Specifically, we provide the LLM with the crafted preceding and subsequent contexts, along with the original text, and ask it to generate a text that 
introduces content variation while preserving essential attributes of the original. 
Therefore, the regenerated text must satisfy the following criteria:
(1) Fitting naturally within the surrounding context;
(2) Aligning  with the original text in terms of theme, length, format, and linguistic style, as mismatches in these aspects between training and testing instances are known to degrade downstream performance \cite{rogers2021primer};
(3) Introducing novel elements to enrich content variation, avoiding simple rewording or direct replication.

In this step, the preceding and subsequent texts generated in Section~\ref{ss:BidirectionalCompletion} serve as a bridge, ensuring the newly generated text aligns with the original,
such as theme and label-related information.
Simultaneously, this process enriches and diversifies the content, making the regenerated text a high-quality augmentation of the original. The prompt design for this step is detailed in Appendix \ref{sec:PromptTemplates}.

\begin{figure}[!t]
    \centering
    \includegraphics[width=3in, trim=0 0 0 0 clip]{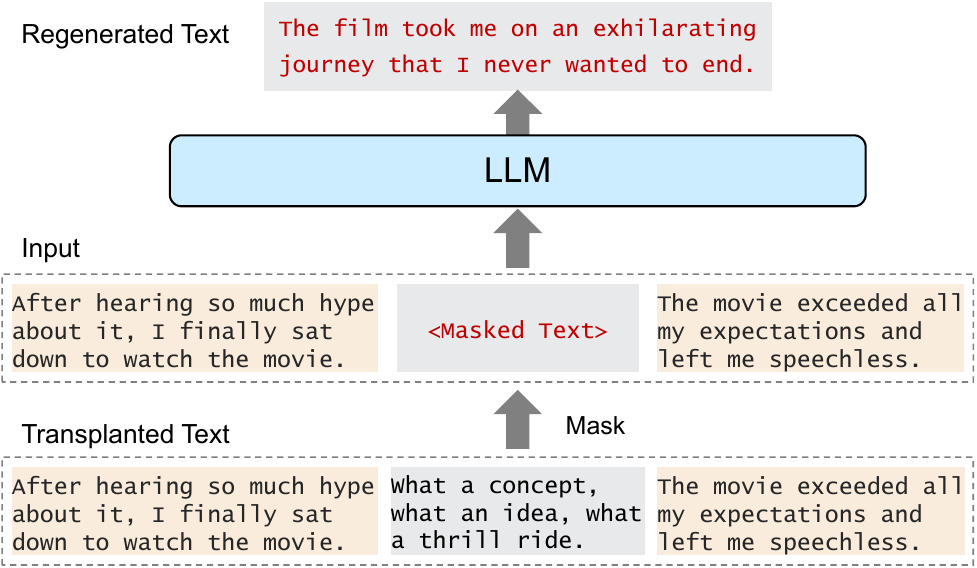}
    \caption{Illustration of regeneration.}
    \label{fig:backinfill}
\end{figure}

\section{Experiments}
We conduct extensive experiments to evaluate the effectiveness of \ourname across multiple deep learning tasks by applying data augmentation to various datasets. Following established practices \cite{yoo-etal-2021-gpt3mix-leveraging, dai2023auggpt, ubani2023zeroshotdataaug, lee2024llm2llm}, we simulate low-resource scenarios by subsampling the training set of each dataset. 
Specifically, we select a subset of samples as seed data for augmentation, and then generate three augmented samples for each seed. 
This enables us to rigorously assess the performance of our method in data-scarce scenarios.

    


\subsection{Implementation Details}

To evaluate \ourname's robustness and effectiveness across a wider range of LLMs, we employ different LLMs as the base LLM for both transplant and regeneration phases, including DeepSeek-V3, GPT-3.5-Turbo and GPT-4o. 


For text classification tasks, we use ModernBERT \cite{ModernBERT}, a modernized bidirectional encoder-only BERT-style model available in two sizes: ModernBERT-base and ModernBERT-large. 
The classifier is initialized using pre-trained models from the Huggingface Transformers library \cite{Huggingface} and optimized with the AdamW optimizer \cite{Adam, 2017Decoupled}. 
We set the learning rate to 4e-5 and maintain other hyperparameters consistent with \citet{fadaee-etal-2017-data}, including 8 training epochs and a batch size of 8. 
During training, models are saved based on their performance on the development set, with the best-performing parameters retained for final evaluation on the test set.

For the question-answering task, we fine-tune Qwen2.5-1.5B \cite{qwen2.5, qwen2} using the AdamW optimizer with a learning rate of 1e-5, a batch size of 8, and 8 training epochs. And for NER task, we fine-tune ModernBERT using the same hyperparameters as in the classification task.

We repeat all experiments 10 times to mitigate the influence of randomness. Additionally, we conduct pairwise two-sample Wilcoxon Signed-Rank tests \cite{PairwiseWilcoxonSignedRankTest} to compare group medians and assess statistical significant differences.
When $p < 0.01$, it indicates that there is a statistical significant difference between the two groups.


\subsection{Datasets}
\label{ss:Datasets}

\begin{table}[]
\centering
\small
\setlength{\tabcolsep}{3pt} 

\begin{tabular}{ll@{}c@{}ccc}
\toprule
\bf Task Type & \bf Dataset\; & \bf Classes\; &  \bf Train & \bf Dev & \bf Test \\ 
\midrule

\multirow{3}{*}{Classification} & SST-2 & 2 & 6228 & 692 & 1821 \\ 
                                & TREC & 6 & 5406 & 546 & 500 \\ 
                                & SNIPS & 7  & 13084 & 700 & 700  \\ 
\hdashline[1pt/2pt]
QA & MLQA & - & 1314 & 437 & 437 \\

\hdashline[1pt/2pt]
NER & CoNLL-2003 & 9 & 3234 & 748 & 679 \\

\bottomrule
\end{tabular}
\caption{Statistics of datasets.}
\label{tab:dataset}

\end{table}

We evaluate \ourname on five established benchmarks:

\begin{itemize}
\setlength{\itemsep}{0pt}
\setlength{\topsep}{0pt}
\setlength{\parsep}{0pt}

\item\textbf{SST-2} \cite{sst2}: 
A widely used 
sentiment classification dataset of movie reviews, 
labeled as ``positive'' or ``negative''. We use the version provided by \citet{ConditionalBERT}\footnote{https://github.com/1024er/cbert\_aug}, which contains 6,228 training samples, 692 development samples, and 1,821 test samples.

\item\textbf{TREC} \cite{trec}: 
A question classification dataset 
annotated with six types: ``Abbreviation'', ``Entity'', ``Description'', ``Human'', ``Location'', and ``Numeric''. Similar to SST-2, we use the version from \citet{ConditionalBERT}, with 5,406 training samples, 546 development samples, and 500 test samples.

\item\textbf{SNIPS} \cite{snips}: 
A text classification dataset 
annotated with seven human intents: ``AddToPlaylist'', ``BookRestaurant'', ``GetWeather'', ``PlayMusic'', ``RateBook'', ``SearchCreativeWork'', and ``SearchScreeningEvent''. We use the version from SlotGated-SLU \cite{goo-etal-2018-slot}\footnote{https://github.com/MiuLab/SlotGated-SLU/tree/master/data/snips}, comprising 13,084 training samples, 700 development samples, and 700 test samples.

\item\textbf{MLQA} \cite{lewis2019mlqa}: 
A 
question-answering benchmark with context passages, questions, and 
answers. 
We use English samples form Huggingface\footnote{https://huggingface.co/datasets/dkoterwa/mlqa\_filtered} and filter out those exceeding 80 tokens to keep a modest length. 

\item\textbf{CoNLL-2003} \cite{CoNLL-2003}: A NER dataset contains four entity types: persons, organizations, locations, and miscellaneous names, tagged using the IOB scheme, resulting in nine distinct IOB labels.

\end{itemize}
Statistics of datasets are summarized in Table \ref{tab:dataset}.

\begin{table*}[ht]
\centering
\small
\setlength{\tabcolsep}{2pt} 
\begin{threeparttable}
\begin{tabular}{lcccccccccc}
\toprule

\bf Method & \multicolumn{2}{c}{\bf SST-2} & \multicolumn{2}{c}{\bf TREC} & \multicolumn{2}{c}{\bf SNIPS} & \multicolumn{2}{c}{\bf MLQA}  & \multicolumn{2}{c}{\bf CoNLL-2003} \\

& \bf Dist-3$\uparrow$ & \bf SV$^*$$\uparrow$ 
& \bf Dist-3$\uparrow$ & \bf SV$^*$$\uparrow$ 
& \bf Dist-3$\uparrow$ & \bf SV$^*$$\uparrow$ 
& \bf Dist-3$\uparrow$ & \bf SV$^*$$\uparrow$ 
& \bf Dist-3$\uparrow$ & \bf SV$^*$$\uparrow$ \\
\hline

Original & 0.99$_{\pm0.01}$ & - & 0.94$_{\pm0.02}$ & - & 0.89$_{\pm0.03}$ & - & 0.97$_{\pm0.01}$ & - & 0.97$_{\pm0.01}$ & - \\
MoreData & 0.99$_{\pm0.01}$ & - & 0.89$_{\pm0.01}$ & - & 0.80$_{\pm0.02}$ & - & 0.93$_{\pm0.02}$ & - & 0.94$_{\pm0.01}$ & -\\

\hdashline[1pt/2pt]

EDA & 0.31$_{\pm0.02}$ & 0.19$_{\pm0.02}$ & 0.46$_{\pm0.02}$ & \underline{0.23}$_{\pm0.01}$ & 0.30$_{\pm0.01}$ & 0.16$_{\pm0.01}$ & 0.40$_{\pm0.01}$ & \underline{0.21}$_{\pm0.01}$  & 0.43$_{\pm0.01}$ & \underline{0.23}$_{\pm0.01}$\\
BackTrans. & 0.69$_{\pm0.03}$ & 0.20$_{\pm0.01}$ & \underline{0.55}$_{\pm0.03}$ & 0.13$_{\pm0.01}$ & \underline{0.60}$_{\pm0.02}$ & 0.19$_{\pm0.01}$ & 0.53$_{\pm0.01}$ & 0.16$_{\pm0.06}$ & 0.62$_{\pm0.03}$ & 0.12$_{\pm0.01}$\\

GPT3Mix & 0.67$_{\pm0.03}$ & - & 0.54$_{\pm0.02}$ & - & 0.50$_{\pm0.04}$ & - & \underline{0.65}$_{\pm0.04}$ & - & 0.18$_{\pm0.02}$ & - \\

AugGPT  & 0.54$_{\pm0.03}$ & \underline{0.24}$_{\pm0.01}$ & 0.37$_{\pm0.02}$ & 0.19$_{\pm0.01}$ & 0.41$_{\pm0.02}$ & \underline{0.27}$_{\pm0.01}$ & 0.38$_{\pm0.02}$ & 0.19$_{\pm0.01}$ & 0.33$_{\pm0.02}$ & 0.22$_{\pm0.01}$ \\
LLM2LLM & \underline{0.71}$_{\pm0.02}$ & 0.23$_{\pm0.01}$ & 0.52$_{\pm0.03}$ & 0.17$_{\pm0.01}$ & 0.54$_{\pm0.01}$ & 0.22$_{\pm0.01}$ & 0.61$_{\pm0.01}$ & 0.17$_{\pm0.01}$ & \underline{0.71}$_{\pm0.02}$ & 0.20$_{\pm0.02}$ \\

\hdashline[1pt/2pt]
\multicolumn{2}{l}{\ourname (ours)} \\

\ (left, right) & \bf 0.88$_{\pm0.03}$ & \bf 0.39$_{\pm0.01}$ & \bf 0.66$_{\pm0.03}$ & \bf 0.30$_{\pm0.01}$ & \bf 0.63$_{\pm0.02}$ & \bf 0.36$_{\pm0.01}$ & \bf 0.72$_{\pm0.03}$ & \bf 0.29$_{\pm0.01}$ & \bf 0.78$_{\pm0.03}$ & \bf 0.27$_{\pm0.01}$\\
\ (right, left) & \bf 0.88$_{\pm0.03}$ & \bf 0.39$_{\pm0.01}$ & 0.64$_{\pm0.03}$ & 0.29$_{\pm0.01}$ & \bf 0.63$_{\pm0.02}$ & \bf 0.36$_{\pm0.01}$ & 0.71$_{\pm0.02}$ & 0.28$_{\pm0.01}$ & 0.75$_{\pm0.03}$ & \bf 0.27$_{\pm0.02}$\\
\ Unidirectional & 0.82$_{\pm0.03}$ & 0.37$_{\pm0.01}$ & 0.54$_{\pm0.02}$ & 0.25$_{\pm0.01}$ & 0.50$_{\pm0.02}$ & 0.31$_{\pm0.01}$ & 0.63$_{\pm0.02}$ & 0.25$_{\pm0.01}$ & 0.66$_{\pm0.03}$ & 0.22$_{\pm0.01}$\\

\bottomrule

\end{tabular}
\end{threeparttable}
\caption{Quality of generated samples by various methods ($p<0.01$). Subscript numbers denote standard deviations. SV $=$ Semantic Variability; MoreData: randomly samples additional data from the original training set as augmented data. 
Underlines indicate the second-best performance for each metric.
We choose DeepSeek-V3 as the base LLM for data augmentation. Results based on GPT-3.5-Turbo and GPT-4o are available in Appendix \ref{sec:IntrinsicEvaluation}.
}
\label{tab:rq1}

\end{table*}

\subsection{Metrics}
We evaluate \ourname along two dimensions: the quality of the augmented texts (intrinsic evaluation) and their impact on deep learning tasks (extrinsic evaluation). For intrinsic evaluation, we adopt two widely employed metrics to assess the quality of the augmented samples: 

\textbf{Distinct-N} measures the lexical diversity~\cite{DISTINCT}. It is defined as the ratio of unique n-grams across all generated texts and their corresponding seeds.

    \begin{align}
        \text{Distinct-N} = \frac{\# \text{ unique n-grams}}{\# \text{ all n-grams}}
    \end{align}
A higher value indicates greater diversity. We calculate the average score across 10 experiments as the final result.

\textbf{Semantic Variability} measures how well the generated text extends the semantics of the seed texts. 
To assess this, we first adopt BERTScore \cite{BERTScore} to calculate the sentence-level similarity between the generated text and its original, leveraging BERT's contextual embeddings. We then define semantic variability as:
    \begin{align}
        \text{Semantic Variability} = 1 - \text{BERTScore}
    \end{align}
Thus, the higher the semantic variability, the better the variability that the new text variants are.

For extrinsic evaluation, we adopt task-related metrics. Specifically, we use accuracy (Acc) and macro F1-score (Macro-F1) for classification 
and NER tasks, and the accuracy of answers (Acc) for QA tasks:
    \begin{align}
        \text{Acc} = \frac{\# \text{samples answered correctly}}{\# \text{all test samples}}
    \end{align}
    
    \begin{equation}
        \text{Macro-F1} = \frac{1}{N} \sum_{i=1}^N F1_i
    \end{equation}
where $N$ is the total number of classes, and $F1_i$ represents the F1-score for the i-th class.

\subsection{Baselines}
\label{ss:Baselines}

We compare our method with both traditional and LLM-based methods:


\noindent\textbf{Easy Data Augmentation (EDA)} \cite{wei-zou-2019-eda}: 
A rule-based data augmentation method that applies lexical transformations, including synonym replacement, random insertion, random swap, and random deletion, to the original text.

\noindent\textbf{Back Translation (BackTrans.)} \cite{2015Improving}: 
A widely used date augmentation method that translates the original text into another language and then back-translates it into the original language to generate variants. Following ZeroShotDataAug \cite{ubani2023zeroshotdataaug}, we use \textit{googletrans}\footnote{Google Translate (https://pypi.org/project/googletrans/)} as the machine translation model, selecting different intermediate languages for multiple augmentations of the same text.

\noindent\textbf{GPT3Mix} \cite{yoo-etal-2021-gpt3mix-leveraging}: 
This method randomly selects several examples from the seed samples, embeds them into a prompt template, and then leverages the powerful few-shot learning capabilities of LLMs to generate mixed augmented text influenced by the provided examples.

\noindent\textbf{AugGPT} \cite{dai2023auggpt}: 
This method utilizes prompts to guide LLMs in rephrasing each sentence from the training samples into multiple 
semantically similar but linguistically different variants, 
thereby enhancing text classification performance.

\noindent\textbf{LLM2LLM} \cite{lee2024llm2llm}: 
An iterative data augmentation strategy that continuously employs LLMs to generate new samples from instances mispredicted by the downstream task model.

\begin{table*}[]
\centering
\small
\setlength{\tabcolsep}{1.2pt} 
\begin{threeparttable}
\begin{tabular}{lccccccccc}
\toprule

\bf Method & \multicolumn{2}{c}{\bf SST-2} & \multicolumn{2}{c}{\bf TREC} & \multicolumn{2}{c}{\bf SNIPS} & \bf MLQA & \multicolumn{2}{c}{\bf CoNLL-2003}\\

& \bf Acc$\uparrow$ & \bf Macro-F1$\uparrow$ & \bf Acc$\uparrow$ & \bf Macro-F1$\uparrow$ & \bf  Acc$\uparrow$ & \bf Macro-F1$\uparrow$ & \bf Acc$\uparrow$ & \bf Acc$\uparrow$ & \bf Macro-F1$\uparrow$ \\
\hline

Original & 52.34$_{\pm3.19}$ & 48.88$_{\pm6.59}$ & 50.80$_{\pm10.60}$ & 47.66$_{\pm9.06}$ & 78.10$_{\pm2.77}$ & 78.30$_{\pm2.68}$ & 32.08$_{\pm1.02}$ & 82.41$_{\pm1.31}$ & 82.44$_{\pm1.14}$ \\
MoreData & 65.41$_{\pm5.48}$ & 65.41$_{\pm5.48}$ & 74.30$_{\pm6.55}$ & 70.75$_{\pm6.47}$ & 88.23$_{\pm3.14}$ & 88.30$_{\pm3.24}$ & 43.30$_{\pm1.13}$ & 91.56$_{\pm0.65}$ & 91.50$_{\pm0.89}$ \\

\hdashline[1pt/2pt]

EDA & 56.78$_{\pm5.04}$ & 55.79$_{\pm5.86}$ & 53.50$_{\pm10.59}$ & 50.58$_{\pm8.76}$ & 81.86$_{\pm4.03}$ & 81.80$_{\pm4.31}$ & 36.92$_{\pm1.67}$ & 85.08$_{\pm0.91}$ & 83.89$_{\pm0.80}$ \\
BackTrans. & 60.09$_{\pm6.25}$ & 58.80$_{\pm8.16}$ & 56.52$_{\pm6.57}$ & 52.49$_{\pm6.37}$ & 81.20$_{\pm3.86}$ & 81.40$_{\pm3.49}$ & 35.32$_{\pm1.24}$ & 85.88$_{\pm1.01}$ & 84.99$_{\pm1.07}$ \\

GPT3Mix & 63.75$_{\pm6.76}$ & \underline{63.36}$_{\pm7.05}$ & 57.68$_{\pm7.12}$ & 53.39$_{\pm7.61}$ & 81.80$_{\pm5.63}$ & 82.15$_{\pm5.24}$ & 33.18$_{\pm1.80}$ & 87.92$_{\pm0.86}$ & 86.75$_{\pm0.79}$ \\
AugGPT & 61.11$_{\pm6.50}$ & 60.55$_{\pm6.94}$ & \underline{58.94}$_{\pm9.43}$ & \underline{55.89}$_{\pm9.95}$ & 82.73$_{\pm3.53}$ & 82.98$_{\pm3.35}$ & 36.04$_{\pm1.45}$ & 87.27$_{\pm0.82}$ & 86.68$_{\pm0.90}$ \\
LLM2LLM & \underline{63.90}$_{\pm5.34}$ & 63.28$_{\pm5.44}$ & 56.58$_{\pm7.17}$ & 52.70$_{\pm6.76}$ & \underline{83.04}$_{\pm4.17}$ & \underline{83.33}$_{\pm4.14}$ & \underline{37.51}$_{\pm1.67}$ & \underline{88.04}$_{\pm1.08}$ & \underline{87.47}$_{\pm1.16}$ \\

\hdashline[1pt/2pt]

\multicolumn{2}{l}{\ourname (ours)} \\

\ (left, right) & \bf 67.08$_{\pm6.92}$ & \bf 66.77$_{\pm6.92}$ & 60.88$_{\pm7.06}$ & 56.38$_{\pm6.73}$ & 84.06$_{\pm2.97}$ & 84.24$_{\pm2.82}$ & 39.44$_{\pm1.54}$ & 90.28$_{\pm0.95}$ & 88.82$_{\pm1.00}$ \\
\ (right, left) & 66.21$_{\pm6.19}$ & 65.34$_{\pm6.43}$ & \bf 60.94$_{\pm8.59}$ & \bf 57.32$_{\pm7.60}$ & \bf 84.10$_{\pm2.82}$ & \bf 84.30$_{\pm2.65}$ & \bf 39.67$_{\pm1.78}$ & \bf 90.86$_{\pm0.90}$ & \bf 89.07$_{\pm1.09}$ \\
\ Unidirectional & 64.34$_{\pm5.38}$ & 63.52$_{\pm5.45}$ & 58.42$_{\pm7.84}$ & 55.06$_{\pm7.05}$ & 82.53$_{\pm4.58}$ & 82.78$_{\pm4.65}$ & 35.69$_{\pm1.67}$ & 87.04$_{\pm0.92}$ & 85.96$_{\pm1.06}$ \\

\bottomrule

\end{tabular}
\end{threeparttable}
\caption{Effectiveness of \ourname in empowering deep learning tasks ($p<$0.01). Subscript numbers denote standard deviations. We choose ModernBERT-base for classification and NER tasks and  Qwen2.5-1.5B for QA task. Detailed results based on ModernBERT-large are available in Appendix \ref{sec:ExtrinsicEvaluation}, Table \ref{tab:ExtrinsicEvaluation_ModernBERT_large}.}
\label{tab:rq2}

\end{table*}

\subsection{Results}
\label{ss:Results}
\subsubsection{Intrinsic Evaluation}
The intrinsic evaluation results in Table \ref{tab:rq1} 
demonstrate that, the quality of augmented samples generated by \ourname significantly outperform other baselines across all benchmarks. In particular, \ourname achieves lexical diversity (Distinct-3) closer to original texts without augmentation (Original) and 
sampling additional data from original training set as augmented data (MoreData), highlighting its effectiveness in improving lexical diversity. Meanwhile, \ourname also exhibits superior semantic variability 
compared to other baselines, indicating that it meaningfully expands the semantics of the original text rather than merely relying on simple rephrasing. 
Additionally, \ourname with bidirectional text continuation outperforms its unidirectional counterpart, which serves as an ablation model for the bidirectional continuation strategy. Further ablation study details are discussed in Section \ref{ss:AblationStudy}.


\subsubsection{Extrinsic Evaluation}
\label{sss:ExtrinsicEvaluation}


We evaluate the effectiveness of \ourname in empowering deep learning tasks by training models on augmented data.
To simulate low-data scenarios, we adopt a subsampling strategy aligned with prior studies \cite{kumar-etal-2020-data, ubani2023zeroshotdataaug}. Specifically, we randomly sample subsets (10 samples per class for classification tasks and 50 samples for QA and NER tasks) from the original training and development sets for each task. These training subsets are expanded using various augmentation methods, including \ourname. The augmented data, combined with the original sub-training set, are used to train deep learning models, and models' performance is evaluated on the original test set.

Results in Table \ref{tab:rq2} demonstrate that the augmented samples generated by \ourname significantly improve task performance. For instance, on the SST-2 dataset, \ourname increases classification accuracy from 52.34\% to 67.08\% and Macro F1-score from 48.88\% to 66.77\%, outperforming even MoreData.
In the QA task on MLQA, \ourname achieves a 23.66\% improvement in accuracy, significantly surpassing other LLM-based baselines.
These findings indicate that, unlike methods that merely prompt LLMs, the transplanting mechanism in \ourname harnesses knowledge embedded in LLMs more effectively, thereby generating higher-quality augmented samples.

\textbf{Time Efficiency}: 
We also compare the time efficiency of different methods, which is defined as the average time required to generate a new sample (Table \ref{tab:TimeEfficiency}). 
To ensure fairness, all comparisons are conducted on the same hardware. 
EDA significantly outperforms other methods, due to its extremely simple rule-based operations. 
Among LLM-based methods, AugGPT is the fastest, attributed to its shortest prompts. 
\ourname is slower but still surpasses GPT3Mix and LLM2LLM. 
LLM2LLM exhibits the lowest efficiency, as its iterative generation and retraining process significantly increases time consumption.




\begin{table}[]
\centering
\small
\begin{threeparttable}

\begin{tabular}{ll}
\toprule
\bf Method & \bf Time   \\ 
\hline
EDA & 0.02$_{\pm0.84}$ \\
BackTrans. & 3.44$_{\pm5.53}$   \\
GPT3Mix & 22.78$_{\pm12.73}$ \\
AugGPT & 4.93$_{\pm8.27}$  \\
LLM2LLM & 30.55$_{\pm11.19}$  \\
\ourname (ours) & 15.09$_{\pm13.87}$ \\
\bottomrule
\end{tabular}
\end{threeparttable}
\caption{Time efficiency for various methods. We report the average processing time (in seconds) required to generate a new sample.}
\label{tab:TimeEfficiency}

\end{table}

\subsubsection{Ablation Study}
\label{ss:AblationStudy}

We hereby conduct ablation experiments to explore the key step in our approach---bidirectional text continuation.
First, we alter the bidirectional text continuation strategy in the transplant step. Specifically, we remove backward continuation and retain only forward continuation (Unidirectional). We also experiment with altering the order of backward and forward continuations, testing both backward-first (left, right) and forward-first (right, left) approaches.

As shown in Table \ref{tab:rq2}, unidirectional continuation significantly reduces effectiveness compared to bidirectional continuation, likely due to the limited context information it provides, resulting in less variation and diversity in the generated text. This is further supported by the obviously lower Distinct-3 score for unidirectional continuation in Table \ref{tab:rq1}. Additionally, the order of bidirectional continuation also affects performance:
forward-first continuation
yields better results on most datasets. This is likely because, compared to backward continuation, LLMs are more proficient in forward continuation.
Specifically, generating subsequent context first provides more information for LLMs to generate a higher-quality preceding context and thereby improve the quality of the augmented data.

\subsubsection{Scaling Analysis}

\begin{figure}[!t] 
\centering
    \begin{subfigure}[b]{\linewidth}
	\includegraphics[scale=0.45, trim=10 0 10 10 clip]{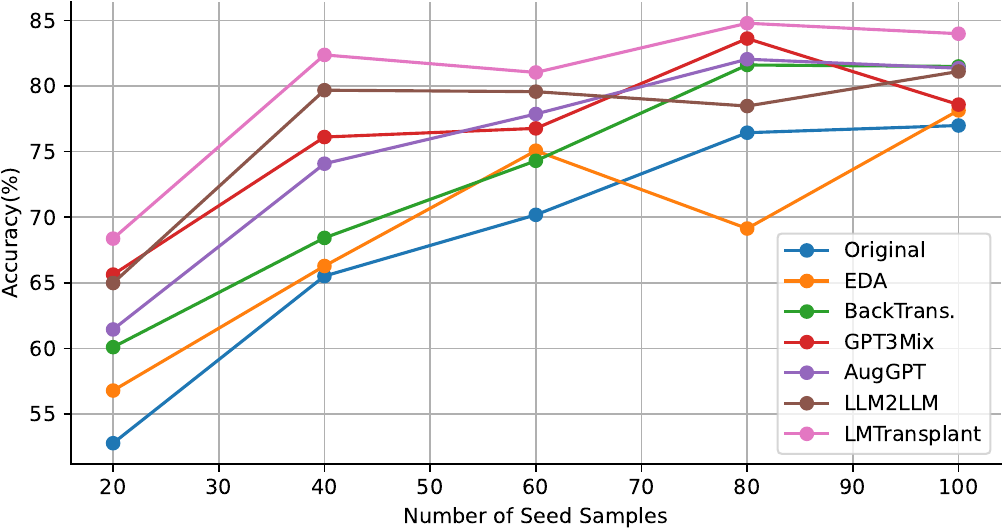}
	\caption{Performance under different numbers of seed samples.}
    \label{seed_num} 
    \end{subfigure}
    \hfill
    \begin{subfigure}[b]{\linewidth}
	\includegraphics[scale=0.45, trim=10 0 10 0 clip]{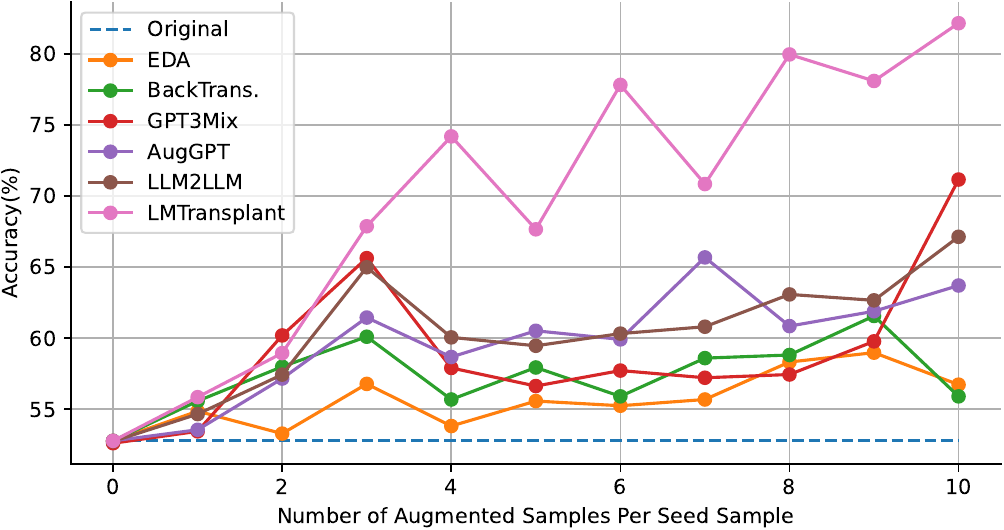}
	\caption{Performance under different numbers of augmented samples per seed sample.}
	\label{aug_num} 
    \end{subfigure}
    \caption{Results of scaling analysis under various sample sizes.}
\label{fig:ablation}
\end{figure}

We further investigate the scaling effect of various sample sizes, including the number of seed samples and augmented samples.
Experiments are conducted on the SST-2 dataset.
First, with the number of augmented samples fixed at 3 per seed, we gradually vary the number of seed samples from 20 to 100 (10 to 50 per category).
Next, with the total number of seed samples fixed at 20 (10 per category), we adjust the number of augmented samples per seed from 0 to 10. After each adjustment, models are retrained and evaluated.
Results illustrated in Figures \ref{fig:ablation} demonstrate that, \ourname consistently achieves the highest accuracy across varying seed sample sizes (Figure~\ref{seed_num}), showcasing its robust generalization under data scarcity. Notably, while other methods plateau after a few rounds of augmentation, \ourname continues to improve with additional augmented samples (Figure~\ref{aug_num}), attributed to its effective utilization of knowledge embedded in LLMs. 
This underscores the exceptional scalability of \ourname and its efficacy in tackling data scarcity challenges.

\subsubsection{Case Study}

\begin{table*} [!h]
\small
\centering
\begin{tabular}{rll}
\toprule
 \textbf{Original Text}: & How big is our galaxy in diameter? & Numeric\\ 
  \hline\hline
  \bf EDA: & How big is our galaxy diameter in & Numeric \\
  \bf BackTrans.: & How great is our galaxy in diameter? & Numeric \\ 
  \bf AugGPT: & What is the diameter of our galaxy? & Numeric \\
  \bf \ourname:  & \textcolor{gray!70}{Astronomers have long been fascinated by the size of our galaxy.} What & \\ 
           & is the approximate number of stars within the Milky Way? \textcolor{gray!70}{The diameter}& Numeric\\
           & \textcolor{gray!70}{of the Milky Way is estimated to be about 100,000 light-years.}& \\
           
\bottomrule
\end{tabular}

\begin{tabular}{rll}
\toprule
\textbf{Original Text}: & What a concept, what an idea, what a thrill ride.  & Positive  \\ 
\hline\hline
\textbf{EDA}: & What a concept what an thought what a thrill ride & Positive \\ 
\textbf{BackTrans.}: & What a concept, what an idea, what alarming ride. & Positive \\ 
\textbf{AugGPT}:  & What a concept, what an idea, what an exhilarating experience. & Positive \\ 
\textbf{\ourname}: & \textcolor{gray!70}{After hearing so much hype about it, I finally sat down to watch the}   &  \\ 

           & \textcolor{gray!70}{movie.} The film took me on an exhilarating journey that I never wanted & Positive \\
           
           & to end. \textcolor{gray!70}{The movie exceeded all my expectations and left me speechless.} & \\

\bottomrule
\end{tabular}
\caption{Cases of augmented samples generated by various methods. The last column shows the classification label.}
\label{table:case}
\end{table*}

We further conduct a qualitative analysis of the augmented samples generated by \ourname. We highlight two cases in 
Table \ref{table:case}, where \ourname extends the diversity and creativity of the original samples while preserving the usability of the augmented samples.


In the first case, the original text is a question asking about the size of our galaxy. We can observe that baseline methods simply rephrase the original text with the same semantics. 
For example, EDA swaps the words ``in'' and ``diameter'', which even disrupts the linguistic integrity of the text. 
Back-translation and AugGPT rephrase the original text by employing machine translation models and LLMs, respectively. While they introduce changes in sentence structure, the new text remains semantically similar to the original.
In contrast, \ourname generates more diverse and creative text centered on the topic of ``our galaxy'', by leveraging the knowledge embedded in LLMs. It expands the phrase ``our galaxy in diameter'' to ``the size of our galaxy'' in the preceding context. More surprisingly, it introduces a new term ``Milky Way'' to represent ``our galaxy'' within the same context. 
This example demonstrates the extraordinary creativity of \ourname in text augmentation.



In the second case, the original text is a movie review expressing the audience's appreciation on specific aspects of a movie. 
Despite the monotonous semantics, we note that the texts generated by baselines exhibit sentiment divergence from the original.
For example, EDA replaces ``idea'' with ``thought'', slightly weakening the emotional intensity of the original text. Back-translation introduces the expression ``alarming ride'' during the translation process, shifting the sentiment from positive to negative or neutral. And AugGPT uses a more abstract expression, ``exhilarating experience''. 
In contrast, \ourname takes the gist of the original text and regenerates a novel review, expressing the same feelings on the theme of ``movie''. The subsequent text retains the positive sentiment with the phrase ``exceeded all my expectations and left me speechless''. The regenerated movie review aligns with the original text both thematically and sentimentally while bringing new knowledge about sentiment classification.

Overall, \ourname effectively harnesses the knowledge embedded in LLMs through its TTR strategy, generating text with enhanced diversity and exceptional creativity that surpasses other augmentation methods. Simultaneously, by utilizing bidirectional context as a bridge, \ourname ensures the new text accurately retains the core characteristics of the original, guaranteeing its usability for downstream tasks.


\section{Discussion}

One concern is that \ourname might generate samples with semantic variations that differ from the original data (as illustrated in Table~\ref{table:case}). This could potentially disrupt the performance of deep learning models, raising questions about the practical utility of the augmented samples.

However, our approach adheres to the fundamental principles of data augmentation: producing diverse, high-quality samples while preserving the original data distribution. By leveraging the bidirectional text continuation process, our method harnesses the knowledge embedded in LLMs to generate more creative texts. This not only enhances the diversity of the data but also improves the generalization capabilities of deep learning models. This hypothesis is supported by the results presented in Section \ref{sss:ExtrinsicEvaluation}, which demonstrate that the texts generated by \ourname significantly boost the performance of deep learning tasks.

Additionally, while the generated texts may exhibit semantic differences from the original, the transplanting mechanism in our approach ensures that the generated samples preserve the core attributes of the seed text---such as theme, linguistic style, and sentiment polarity. This mechanism effectively mitigates the risk of producing nonsensical or irrelevant text samples, thereby maintaining the integrity and usefulness of the augmented data.

\section{Related Work}
Data augmentation has been extensively explored in NLP tasks \cite{feng-etal-2021-survey, 2023-survey, chen-etal-2023-survey}. Popular approaches such as word-level transformations \cite{wei-zou-2019-eda} often disrupt semantic coherence, while Back-translation \cite{2015Improving} offers limited diversity due to high similarity with the source text. In contrast, \ourname introduces a novel paradigm for text data augmentation, generating content-level variants that are more diverse and creative in content.

Recently, LLM-based data augmentation has gained wider attention \cite{whitehouse-etal-2023-llm, ding-etal-2024-data, zhou2024survey, qiao2024autoact}. Leveraging their ``knowledge emergence'' capability, LLMs can generate desired content directly through demonstrations or natural language instructions. 
For instance, AugGPT \cite{dai2023auggpt} uses natural language instructions to guide ChatGPT in rephrasing the original text. However, this approach applies relatively simple operations and underutilizes the extensive knowledge embedded in LLMs, resulting in limited content creativity and diversity. In comparison, \ourname simulates realistic contextual scenarios, enabling LLMs to better leverage their knowledge and generate more diverse and creative augmented samples.

Similarly, GPT3Mix \cite{yoo-etal-2021-gpt3mix-leveraging} leverages LLMs' few-shot learning capabilities by providing a set of examples to generate new samples. However, it is highly sensitive to example quality, selection strategy, and example order, potentially introducing uncontrolled biases that compromise augmentation stability. More recently, LLM2LLM \cite{lee2024llm2llm} iteratively augments instances misclassified by downstream-task model. While this approach produces more targeted samples, its iterative generation and retraining process incurs high computational costs and lacks adaptability to new datasets. In contrast, \ourname utilizes LLMs' powerful language understanding and instruction-following abilities through carefully designed prompts, eliminating the need for examples. This approach offers greater flexibility for adapting to different downstream tasks while mitigating issues related to data selection sensitivity and excessive computational resource consumption.


\section{Conclusion}

In this paper, we propose \ourname, a novel text data augmentation paradigm based on transplanting strategy. By leveraging bidirectional text continuation and masked text prediction, \ourname generates high-quality and diverse augmented text. It constructs contextually coherent scenarios aligned with the original text, fully utilizing the knowledge embedded in LLMs. Experimental results demonstrate that the augmented text generated by \ourname excels in diversity and creativity while significantly improving the performance of downstream tasks.


Our replication package is available 
at: ~\href{https://github.com/W-GZ/LMTransplant}{https://github.com/W-GZ/LMTransplant}.


\section*{Limitations}

Although \ourname demonstrates strong experimental results, we acknowledge the following limitations and challenges that warrant further investigation: 
First, our experiments are conducted based on DeepSeek-V3, GPT-3.5-Turbo, and GPT-4o. We plan to extend our evaluation across a wider range of LLMs to further assess the robustness and effectiveness of \ourname.
Second, in line with prior LLM-based augmentation methods (e.g., GPT3Mix, LLM2LLM), we evaluate \ourname on classification, question-answering and named entity recognition tasks, which we believe are representative tasks to demonstrate the generalization capability of \ourname. However, when applying \ourname to other task types, appropriate adjustments to its prompts may be necessary to ensure adaptability and effectiveness. As such, we plan to investigate its broader applicability in future work.



\section*{Acknowledgment}
This research is funded by the National Key Research and Development Program of China (Grant No. 2023YFB4503802) and the Natural Science Foundation of Shanghai (Grant No. 25ZR1401175).

\bibliography{ref}


\section*{Appendix}
\appendix

\begin{table*}[ht]
\centering
\small
\setlength{\tabcolsep}{2pt} 
\begin{threeparttable}
\begin{tabular}{lcccccccccc}
\toprule

\bf Method & \multicolumn{2}{c}{\bf SST-2} & \multicolumn{2}{c}{\bf TREC} & \multicolumn{2}{c}{\bf SNIPS} & \multicolumn{2}{c}{\bf MLQA}  & \multicolumn{2}{c}{\bf CoNLL-2003} \\

& \bf Dist-3$\uparrow$ & \bf SV$^*$$\uparrow$ 
& \bf Dist-3$\uparrow$ & \bf SV$^*$$\uparrow$ 
& \bf Dist-3$\uparrow$ & \bf SV$^*$$\uparrow$ 
& \bf Dist-3$\uparrow$ & \bf SV$^*$$\uparrow$ 
& \bf Dist-3$\uparrow$ & \bf SV$^*$$\uparrow$ \\
\hline

Original & 0.99$_{\pm0.01}$ & - & 0.94$_{\pm0.02}$ & - & 0.89$_{\pm0.03}$ & - & 0.97$_{\pm0.01}$ & - & 0.97$_{\pm0.01}$ & - \\
MoreData & 0.99$_{\pm0.01}$ & - & 0.89$_{\pm0.01}$ & - & 0.80$_{\pm0.02}$ & - & 0.93$_{\pm0.02}$ & - & 0.94$_{\pm0.01}$ & -\\

\hdashline[1pt/2pt]

EDA & 0.31$_{\pm0.02}$ & 0.19$_{\pm0.02}$ & 0.46$_{\pm0.02}$ & \underline{0.23}$_{\pm0.01}$ & 0.30$_{\pm0.01}$ & 0.16$_{\pm0.01}$ & 0.40$_{\pm0.01}$ & \underline{0.21}$_{\pm0.01}$ & 0.43$_{\pm0.01}$ & \underline{0.23}$_{\pm0.01}$\\
BackTrans. & \underline{0.69}$_{\pm0.03}$ & 0.20$_{\pm0.01}$ & \underline{0.55}$_{\pm0.03}$ & 0.13$_{\pm0.01}$ & \underline{0.60}$_{\pm0.02}$ & 0.19$_{\pm0.01}$ & 0.53$_{\pm0.01}$ & 0.16$_{\pm0.06}$ & 0.62$_{\pm0.03}$ & 0.12$_{\pm0.01}$\\

GPT3Mix & 0.65$_{\pm0.02}$ & - & 0.43$_{\pm0.02}$ & - & 0.45$_{\pm0.02}$ & - & \underline{0.61}$_{\pm0.02}$ & - & 0.45$_{\pm0.03}$ & - \\
AugGPT & 0.43$_{\pm0.02}$ & 0.20$_{\pm0.01}$ & 0.28$_{\pm0.01}$ & 0.15$_{\pm0.01}$ & 0.28$_{\pm0.01}$ & \underline{0.23}$_{\pm0.01}$ & 0.32$_{\pm0.01}$ & 0.15$_{\pm0.01}$ & 0.32$_{\pm0.02}$ & 0.17$_{\pm0.01}$ \\
LLM2LLM & 0.67$_{\pm0.02}$ & \underline{0.22}$_{\pm0.02}$ & 0.50$_{\pm0.02}$ & 0.17$_{\pm0.01}$ & 0.49$_{\pm0.01}$ & 0.20$_{\pm0.01}$ & 0.57$_{\pm0.01}$ & 0.16$_{\pm0.01}$ & \underline{0.68}$_{\pm0.03}$ & 0.19$_{\pm0.02}$ \\

\hdashline[1pt/2pt]

\multicolumn{2}{l}{\ourname (ours)} \\

\ (left, right) & \bf 0.86$_{\pm0.02}$ & \bf 0.37$_{\pm0.01}$ & \bf 0.67$_{\pm0.02}$ & \bf 0.28$_{\pm0.01}$ & \bf 0.66$_{\pm0.02}$ & \bf 0.35$_{\pm0.01}$ & \bf 0.70$_{\pm0.01}$ & \bf 0.27$_{\pm0.01}$ & \bf 0.73$_{\pm0.02}$ & \bf 0.24$_{\pm0.01}$ \\
\ (right, left) & 0.85$_{\pm0.02}$ & \bf 0.37$_{\pm0.01}$ & 0.66$_{\pm0.02}$ & 0.27$_{\pm0.01}$ & \bf 0.66$_{\pm0.02}$ & 0.34$_{\pm0.01}$ & 0.69$_{\pm0.02}$ & \bf 0.27$_{\pm0.01}$ & 0.72$_{\pm0.02}$ & \bf 0.24$_{\pm0.01}$ \\
\ Unidirectional & 0.81$_{\pm0.03}$ & 0.35$_{\pm0.01}$ & 0.58$_{\pm0.03}$ & 0.25$_{\pm0.01}$ & 0.57$_{\pm0.02}$ & 0.31$_{\pm0.01}$ & 0.60$_{\pm0.01}$ & 0.24$_{\pm0.01}$ & 0.59$_{\pm0.04}$ & 0.22$_{\pm0.01}$ \\

\bottomrule
\end{tabular}
\end{threeparttable}
\caption{Intrinsic evaluation results based on GPT-3.5-Turbo. Subscript numbers denote standard deviations.}
\label{tab:IntrinsicEvaluation_GPT3.5}
\end{table*}

\begin{table*}[ht]
\centering
\small
\setlength{\tabcolsep}{2pt} 
\begin{threeparttable}
\begin{tabular}{lcccccccccc}
\toprule

\bf Method & \multicolumn{2}{c}{\bf SST-2} & \multicolumn{2}{c}{\bf TREC} & \multicolumn{2}{c}{\bf SNIPS} & \multicolumn{2}{c}{\bf MLQA}  & \multicolumn{2}{c}{\bf CoNLL-2003} \\

& \bf Dist-3$\uparrow$ & \bf SV$^*$$\uparrow$ 
& \bf Dist-3$\uparrow$ & \bf SV$^*$$\uparrow$ 
& \bf Dist-3$\uparrow$ & \bf SV$^*$$\uparrow$ 
& \bf Dist-3$\uparrow$ & \bf SV$^*$$\uparrow$ 
& \bf Dist-3$\uparrow$ & \bf SV$^*$$\uparrow$ \\
\hline

Original & 0.99$_{\pm0.01}$ & - & 0.94$_{\pm0.02}$ & - & 0.89$_{\pm0.03}$ & - & 0.97$_{\pm0.01}$ & - & 0.97$_{\pm0.01}$ & - \\
MoreData & 0.99$_{\pm0.01}$ & - & 0.89$_{\pm0.01}$ & - & 0.80$_{\pm0.02}$ & - & 0.93$_{\pm0.02}$ & - & 0.94$_{\pm0.01}$ & -\\

\hdashline[1pt/2pt]

EDA & 0.31$_{\pm0.02}$ & 0.19$_{\pm0.02}$ & 0.46$_{\pm0.02}$ & \underline{0.23}$_{\pm0.01}$ & 0.30$_{\pm0.01}$ & 0.16$_{\pm0.01}$ & 0.40$_{\pm0.01}$ & 0.21$_{\pm0.01}$ & 0.43$_{\pm0.01}$ & 0.23$_{\pm0.01}$\\
BackTrans. & 0.69$_{\pm0.03}$ & 0.20$_{\pm0.01}$ & 0.55$_{\pm0.03}$ & 0.13$_{\pm0.01}$ & \underline{0.60}$_{\pm0.02}$ & 0.19$_{\pm0.01}$ & 0.53$_{\pm0.01}$ & 0.16$_{\pm0.06}$ & 0.62$_{\pm0.03}$ & 0.12$_{\pm0.01}$\\

GPT3Mix & \underline{0.81}$_{\pm0.02}$ & - & 0.50$_{\pm0.03}$ & - & 0.53$_{\pm0.01}$ & - & \underline{0.70}$_{\pm0.02}$ & - & 0.57$_{\pm0.02}$ & - \\
AugGPT & 0.63$_{\pm0.03}$ & 0.23$_{\pm0.01}$ & 0.40$_{\pm0.03}$ & 0.17$_{\pm0.01}$ & 0.41$_{\pm0.02}$ & \underline{0.26}$_{\pm0.01}$ & 0.54$_{\pm0.03}$ & \underline{0.26}$_{\pm0.01}$ & 0.40$_{\pm0.02}$ & 0.20$_{\pm0.01}$\\
LLM2LLM & 0.79$_{\pm0.02}$ & \underline{0.25}$_{\pm0.01}$ & \underline{0.56}$_{\pm0.02}$ & 0.20$_{\pm0.01}$ & 0.58$_{\pm0.01}$ & 0.23$_{\pm0.01}$ & 0.68$_{\pm0.01}$ & 0.22$_{\pm0.01}$ & \underline{0.79}$_{\pm0.02}$ & \underline{0.24}$_{\pm0.01}$\\

\hdashline[1pt/2pt]

\multicolumn{2}{l}{\ourname (ours)} \\

\ (left, right) & \bf 0.92$_{\pm0.02}$ & \bf 0.46$_{\pm0.01}$ & \bf 0.79$_{\pm0.02}$ & \bf 0.35$_{\pm0.01}$ & \bf 0.78$_{\pm0.01}$ & \bf 0.43$_{\pm0.01}$ & \bf 0.85$_{\pm0.01}$ & \bf 0.38$_{\pm0.01}$ & \bf 0.86$_{\pm0.02}$ & \bf 0.27$_{\pm0.01}$ \\
\ (right, left) & \bf 0.92$_{\pm0.02}$ & 0.45$_{\pm0.02}$ & 0.77$_{\pm0.01}$ & 0.32$_{\pm0.01}$ & \bf 0.78$_{\pm0.02}$ & 0.39$_{\pm0.01}$ & 0.83$_{\pm0.01}$ & 0.37$_{\pm0.01}$ & \bf 0.86$_{\pm0.03}$ & \bf 0.27$_{\pm0.01}$ \\
\ Unidirectional & 0.86$_{\pm0.02}$ & 0.45$_{\pm0.01}$ & 0.67$_{\pm0.02}$ & 0.29$_{\pm0.01}$ & 0.71$_{\pm0.01}$ & 0.36$_{\pm0.01}$ & 0.76$_{\pm0.02}$ & 0.35$_{\pm0.01}$ & 0.76$_{\pm0.02}$ & 0.25$_{\pm0.01}$\\

\bottomrule
\end{tabular}
\end{threeparttable}
\caption{Intrinsic evaluation results based on GPT-4o. Subscript numbers denote standard deviations.}
\label{tab:IntrinsicEvaluation_GPT4o}
\end{table*}

\section{Prompt Design}
\label{sec:PromptTemplates}

Here, we provide the prompts used for transplant and regeneration.

\textbf{Prompt for Transplant}: 
In the transplant process, we treat the original text as a fragment of a broader contextual passage and use LLMs to generate the relevant preceding and subsequent contexts. 
This process can be conceptualized as bidirectional text continuation, which consists of two steps: 
(1) forward continuation (left-to-right), where the model continues writing the subsequent context of the seed text, and (2) backward continuation (right-to-left) , where the model reconstructs the preceding context of the seed text.

Utilizing the powerful language understanding and instruction-following capabilities of LLMs, we employ prompt engineering to guide bidirectional text continuation. Below is the prompt template for our bidirectional text continuation process. First, we instruct LLMs to generate a subsequent text that seamlessly connects to the seed text while maintaining logical coherence. Then, the generated text is combined with the seed text to form an extended passage. Following this, we guide LLMs to generate a preceding text using this expanded input. Through this process, the generated context remains thematically aligned with the original text, while benefiting from enhanced expressiveness and enriched information through the knowledge embedded in LLMs. 

\begin{tcolorbox}[breakable, title=Prompt for Transplant]
\scriptsize
  \texttt{Given the original <text\_type>, generate a subsequent sentence and a preceding sentence as follows:} \\
  \texttt{Subsequent Sentence: Generate a sentence that can naturally follow the original text, ensuring a smooth transition and logical continuation.} \\
  \texttt{New Text: Place the subsequent sentence behind the original <text\_type> to form a new text.} \\
  \texttt{Preceding Sentence: Create a sentence that can naturally precede the new <text\_type>, making the transition smooth and logical.} \\
  
  \texttt{The original <text\_type> is: <original text>} \\
  
  \texttt{Now please return the generated subsequent sentence and preceding sentence in the following format:} \\
  \texttt{Preceding Sentence: [The generated preceding sentence]} \\
  \texttt{Original Text: [The original <text\_type>]} \\
  \texttt{Subsequent Sentence: [The generated subsequent sentence]} \\

\end{tcolorbox}

\textbf{Prompt for Regeneration}: 
The bidirectional text continuation phase generates multiple contextual scenes bearing the original text. However, these transplanted texts cannot be directly used as augmented data. Firstly, they all contain the same original text, which sacrifices the diversity of the augmented data. Secondly, these generated texts exhibit significant differences from the original text in various aspects, such as text length and format, making their differences from the original text too obvious. This is not ideal for data augmentation.

Therefore, we introduce a regeneration process, which uses LLMs to regenerate new text variants that seamlessly integrate into the expanded context. Specifically, we provide LLMs with the crafted preceding and subsequent contexts, along with the original text, and ask it to generate a new text. 
We implement this process using a prompt. The specific prompt template is as follows. In the prompt, we also specify some requirements for the new text generation: first, the new text should connect smoothly with the surrounding context, ensuring semantic consistency and logical flow; second, the new text should align with the original text in length, format, and linguistic style; and finally, the new text must not be a simple modification or direct copy of the original text. Instead, it should introduce novel elements, enhancing the richness and variation of the content.
The prompt also explicitly instructs the LLM to adhere strictly to the original text's label category (denoted as ``<label\_type>'').

\begin{tcolorbox}[breakable, title=Prompt for Regeneration]
\scriptsize
 \texttt{You are provided with three pieces of text:}\\
 \texttt{1. Preceding Sentence: <preceding sentence>}\\
 \texttt{2. Original <text\_type>: <original text>}\\
 \texttt{3. Subsequent Sentence: <subsequent sentence>}\\

 \texttt{You are an expert in text data augmentation. Your task is to generate a new <text\_type> that can replace the original <text\_type> while meeting these requirements:}\\
 \texttt{1. Fits naturally between the preceding sentence and subsequent sentence, maintaining logical flow and coherence.}\\
 \texttt{2. Similar in text length, format (sentence pair, etc.), and language style to the original <text\_type>.}\\
 \texttt{3. Similar '<label\_type>' to the original <text\_type>, which is '<original label>'.}\\
 \texttt{4. The new <text\_type> should not simply repeat the original <text\_type>.}\\

 \texttt{Now please return the generated new <text\_type> as 'Middle Sentence' in the following format:}\\
 \texttt{Preceding Sentence: [The provided preceding sentence]
Middle Sentence: [The generated new <text\_type>]}\\
 \texttt{Subsequent Sentence: [The provided subsequent sentence]}\\

\end{tcolorbox}

\begin{table}[]
\centering
\small
\setlength{\tabcolsep}{2pt} 
\begin{threeparttable}
\begin{tabular}{lccccc}
\toprule

\bf Method & \bf SST-2 & \bf TREC & \bf SNIPS & \bf MLQA & \bf CoNLL \\
\hline

Original & 1.00 & 0.97 & 0.94 & 0.98 & 0.98 \\
MoreData & 0.99 & 0.94 & 0.87 & 0.95 & 0.96 \\
\hdashline[1pt/2pt]

EDA & 0.35 & 0.53 & 0.35 & 0.43 & 0.47 \\
BackTrans. & 0.71 & \underline{0.57} & \underline{0.66} & 0.55 & 0.63 \\
AugGPT & 0.55 & 0.40 & 0.45 & 0.39 & 0.35 \\
LLM2LLM & \underline{0.73} & 0.56 & 0.59 & \underline{0.63} & \underline{0.72} \\
\hdashline[1pt/2pt]

\multicolumn{2}{l}{\ourname (ours)} \\
\ (left, right) & \bf 0.89 & \bf 0.70 & \bf 0.74 & \bf 0.75 & \bf 0.82 \\
\ (right, left) & \bf 0.89 & 0.68 & \bf 0.74 & 0.73 & 0.80 \\
\ Unidirectional & 0.83 & 0.58 & 0.59 & 0.65 & 0.67 \\

\bottomrule
\end{tabular}
\end{threeparttable}
\caption{Compute Distinct-N per seed and average across seeds.}
\label{tab:Distinct-N_per_seed}
\end{table}

\begin{table}[]
\centering
\small
\setlength{\tabcolsep}{1.2pt} 
\begin{threeparttable}
\begin{tabular}{lccccc}
\toprule

\bf Method & \bf SST-2 & \bf TREC & \bf SNIPS & \bf MLQA & \bf CoNLL \\
\hline

Original & 342.60 & 457.80 & 438.50 & 283.30 & 593.20 \\
MoreData & 1375.50 & 1687.90 & 1554.10 & 1117.40 & 2279.30 \\
\hdashline[1pt/2pt]

EDA & 389.50 & 803.60 & 587.50 & 538.40 & 892.80 \\
BackTrans. & 867.30 & 935.10 & \underline{1150.80} & 550.40 & \underline{1340.80} \\
GPT3Mix & 927.80 & \underline{1074.20} & 980.20 & 739.00 & 962.30 \\
AugGPT & 717.90 & 800.80 & 962.30 & 498.70 & 806.90 \\
LLM2LLM & \underline{964.40} & 998.60 & 1070.50 & \underline{791.50} & 1022.21 \\
\hdashline[1pt/2pt]

\multicolumn{2}{l}{\ourname (ours)} \\
\ (left, right) & \bf 1144.00 & \bf 1384.10 & \bf 1490.60 & \bf 927.70 & 1471.30 \\
\ (right, left) & 1129.80 & 1343.20 & 1466.90 & 896.60 & \bf 1534.00 \\
\ Unidirectional & 1113.10 & 1105.10 & 1154.90 & 759.80 & 1172.60 \\

\bottomrule
\end{tabular}
\end{threeparttable}
\caption{Numbers of unique n-grams for various methods.}
\label{tab:Distinct-n-grams}
\end{table}

\section{Intrinsic Evaluation}
\label{sec:IntrinsicEvaluation}
The intrinsic evaluation results based on GPT-3.5-Turbo and GPT-4o are provided in Table \ref{tab:IntrinsicEvaluation_GPT3.5} and Table \ref{tab:IntrinsicEvaluation_GPT4o}, respectively. These results closely align with those obtained based on DeepSeek-V3 (Table \ref{tab:rq1}), demonstrating that \ourname consistently generates high-quality augmentations across different LLM architectures, highlighting its robustness and effectiveness.

We also explore alternative ways of computing Distinct-N and Semantic Variability to further validate the high quality of samples generated by \ourname.

\textbf{Distinct-N}: We calculate Distinct-N across all seeds and their corresponding augmentations. However, repeated n-grams across different seeds may affect the results. To address this, we compute Distinct-N per seed and then average across all seeds. As shown in the Table \ref{tab:Distinct-N_per_seed}, this method generally yields higher scores for all augmentation methods, as it avoids counting repeated n-grams across different seeds. However, since the seeds themselves already exhibit low redundancy in n-grams (with Original's Distinct-N close to 1.0), the overall improvement is limited. This indicates that when the seeds already have low n-gram redundancy, computing Distinct-N per seed or globally makes little difference. Additionally, this method is not applicable to GPT3Mix, as each generated text corresponds to multiple seeds, so GPT3Mix was excluded from this comparison.

The ratio-based Distinct-N metric also has limitations---when seed texts already have high diversity, and augmented samples are diverse but larger in scale, the ratio-based metric may fail to distinguish which augmentation method has more advantages. To better capture this, we additionally compute the number of unique n-grams in the seeds, and in the combined set of seeds and augmented texts. As shown in the Table \ref{tab:Distinct-n-grams}, \ourname consistently achieves significantly higher scores than other methods, approaching those of MoreData, highlighting its ability to introduce a substantial number of novel words and expressions.

\begin{table}[]
\centering
\small
\setlength{\tabcolsep}{2pt} 
\begin{threeparttable}
\begin{tabular}{lccccc}
\toprule

\bf Method & \bf SST-2 & \bf TREC & \bf SNIPS & \bf MLQA & \bf CoNLL \\
\hline

EDA & \underline{0.17} & \underline{0.13} & \underline{0.20} & \underline{0.14} & 0.14 \\
BackTrans. & 0.09 & 0.10 & 0.11 & 0.07 & \underline{0.15} \\
AugGPT & 0.12 & 0.09 & 0.12 & 0.09 & 0.05 \\
LLM2LLM & 0.15 & 0.11 & 0.14 & 0.10 & 0.12 \\
\hdashline[1pt/2pt]

\multicolumn{2}{l}{\ourname (ours)} \\
\ (left, right) & \bf 0.29 & \bf 0.21 & \bf 0.25 & \bf 0.18 & \bf 0.18 \\
\ (right, left) & \bf 0.29 & \bf 0.21 & \bf 0.25 & 0.16 & 0.17 \\
\ Unidirectional & 0.26 & 0.16 & 0.17 & 0.14 & 0.11 \\

\bottomrule
\end{tabular}
\end{threeparttable}
\caption{The differences between samples generated per seed.}
\label{tab:SemanticVariability_per_seed}
\end{table}

\begin{table*}[ht]
\centering
\small
\begin{threeparttable}
\begin{tabular}{lcccccc}
\toprule

\bf Method & \multicolumn{2}{c}{\bf SST-2} & \multicolumn{2}{c}{\bf TREC} & \multicolumn{2}{c}{\bf SNIPS} \\
\cline{2-7}

& \bf Acc$\uparrow$ & \bf Macro-F1$\uparrow$ & \bf Acc$\uparrow$ & \bf Macro-F1$\uparrow$ & \bf  Acc$\uparrow$ & \bf Macro-F1$\uparrow$ \\
\hline

Original & 54.77$_{\pm3.57}$ & 53.97$_{\pm3.97}$ & 51.80$_{\pm7.44}$ & 48.52$_{\pm6.17}$ & 86.17$_{\pm2.88}$ & 86.23$_{\pm2.91}$ \\
MoreData & 71.57$_{\pm3.90}$ & 71.12$_{\pm4.07}$ & 72.08$_{\pm8.11}$ & 68.90$_{\pm8.19}$ & 91.53$_{\pm2.04}$ & 91.67$_{\pm2.08}$ \\

\hdashline[1pt/2pt]

EDA & 59.69$_{\pm3.22}$ & 58.72$_{\pm3.56}$ & 57.22$_{\pm7.33}$ & 54.08$_{\pm8.10}$ & \underline{88.87}$_{\pm2.31}$ & \underline{88.88}$_{\pm2.36}$ \\
BackTrans. & 61.76$_{\pm4.44}$ & 60.78$_{\pm5.41}$ & 55.38$_{\pm8.72}$ & 52.17$_{\pm8.03}$ & 87.97$_{\pm2.93}$ & 88.13$_{\pm2.78}$ \\
GPT3Mix & 66.91$_{\pm2.67}$ & 66.64$_{\pm2.68}$ & 58.52$_{\pm7.49}$ & \underline{56.68}$_{\pm6.32}$ & 88.19$_{\pm3.12}$ & 88.11$_{\pm3.32}$ \\
AugGPT & 63.62$_{\pm4.04}$ & 62.79$_{\pm3.98}$ & 55.82$_{\pm5.57}$ & 53.86$_{\pm5.09}$ & 87.10$_{\pm4.35}$ & 87.36$_{\pm4.20}$ \\
LLM2LLM & \underline{67.38}$_{\pm3.07}$ & \underline{67.33}$_{\pm5.60}$ & \underline{59.56}$_{\pm8.90}$ & 55.95$_{\pm8.45}$ & 88.84$_{\pm1.94}$ & 88.86$_{\pm1.91}$ \\

\hdashline[1pt/2pt]

\multicolumn{2}{l}{\ourname (ours)} \\
\ (left, right) & \bf 71.89$_{\pm3.80}$ & \bf 71.47$_{\pm4.29}$ & 61.18$_{\pm5.70}$ & 59.88$_{\pm5.65}$ & 89.07$_{\pm4.25}$ & 89.08$_{\pm4.66}$ \\
\ (right, left) & 70.46$_{\pm4.89}$ & 69.56$_{\pm5.40}$ & \bf 62.22$_{\pm8.74}$ & \bf 60.37$_{\pm8.52}$ & \bf 89.60$_{\pm3.24}$ & \bf 89.70$_{\pm3.38}$ \\
\ Unidirectional & 68.19$_{\pm3.98}$ & 67.66$_{\pm3.96}$ & 56.46$_{\pm6.07}$ & 54.19$_{\pm7.10}$ & 88.03$_{\pm2.84}$ & 88.23$_{\pm2.87}$ \\

\bottomrule

\end{tabular}
\end{threeparttable}
\caption{Extrinsic evaluation results based on ModernBERT-large. Subscript numbers denote standard deviations.}
\label{tab:ExtrinsicEvaluation_ModernBERT_large}
\end{table*}

\textbf{Semantic Variability}: To capture differences between samples generated per single seed, we conduct an additional experiment: for each seed, we compute pairwise BERTScore among its three augmented texts. We average these pairwise scores per seed and then compute the mean across all seeds. To ensure reliability, we repeat this process over 10 independent trials and report the mean score. Finally, we define 1 - BERTScore as the metric to quantify the difference among augmented texts generated for the same seed. The results can be found in the Table \ref{tab:SemanticVariability_per_seed}.

Results show that \ourname consistently achieves higher variability than other methods, further demonstrating its ability to generate diverse augmentations.

\section{Extrinsic Evaluation}
\label{sec:ExtrinsicEvaluation}

The extrinsic evaluation results of text classification tasks based on ModernBERT-large are provided in Table \ref{tab:ExtrinsicEvaluation_ModernBERT_large}.
Similar to the results based on ModernBERT-base (provided in Table \ref{tab:rq2}), the augmented samples generated by \ourname significantly improve task performance, surpassing other methods across all datasets. Meanwhile, bidirectional continuation outperforms unidirectional continuation, primarily because unidirectional continuation provides more limited contextual information, resulting in less variation and diversity in the generated text.



\section{Label Consistency}
\label{ss:LabelConsistency}
Ensuring the augmented text maintains label consistency is a key factor in determining the quality of augmented samples. \ourname incorporates two key mechanisms to preserve label consistency: (1) The bidirectional context continuation during the transplant phase serves as a bridge between the original and augmented text. This ensures that the augmented text accurately preserves essential attributes of the original, including label-related information. (2) Prompt constraints: in the regeneration phase, we employ explicit prompt constraints (as seen in Appendix \ref{sec:PromptTemplates}) to instruct the LLM to adhere strictly to the original text's label category (denoted as ``<label\_type>'').

\begin{table}[]
\centering
\small
\begin{threeparttable}
\begin{tabular}{lccc}
\toprule

\bf Method & \bf SST-2 & \bf TREC & \bf SNIPS \\

\hline

Original & 0.9500 & 0.6750 & 0.9543 \\
MoreData & 0.9363 & 0.7042 & 0.9325 \\

\hdashline[1pt/2pt]

EDA & 0.9275 & 0.6192 & 0.9100 \\
BackTrans. & 0.8975 & 0.6333 & 0.9118 \\

GPT3Mix & \underline{0.9708} & 0.6414 & \bf 0.9691 \\
AugGPT & 0.9312 & 0.6771 & 0.9429 \\
LLM2LLM & 0.9425 & \underline{0.6916} & 0.9214 \\

\hdashline[1pt/2pt]

\multicolumn{2}{l}{\ourname (ours)} \\

\ (left, right) & 0.9663 & \bf 0.7283 & 0.9161 \\
\ (right, left) & 0.9688 & 0.7154 & 0.9354 \\
\ Unidirectional & \bf 0.9750 & 0.7125 & \underline{0.9439} \\
\bottomrule

\end{tabular}
\end{threeparttable}
\caption{Label consistency for various methods.}
\label{tab:LabelConsistency}

\end{table}

To further validate label consistency, we employ an LLM to predict the labels of augmented texts and compute the ratio of generated labels that match the original labels. The prompt is as follows:

\begin{tcolorbox}[breakable, title=Prompt for Label Prediction]
\scriptsize
  \texttt{You are an expert text classifier. Your task is to analyze the given text and select exactly ONE most appropriate label from the provided candidate label list.} \\
  
  \texttt{Text: \{text\}} \\
  \texttt{Candidate label list: \{self.label\_enum\_str(label\_set)\}} \\
  
  \texttt{Please return ONLY the selected label name without explanations, punctuation or additional text.} \\
\end{tcolorbox}

The results are shown in Table \ref{tab:LabelConsistency}. We notice that \ourname effectively maintains label consistency in augmented texts. This demonstrates that
\ourname can produce diverse and creative samples while preserving the core attributes of the original text, thereby maintaining the integrity and usefulness of the augmented data.

\section{Generate IOB Label for NER Task}

For named entity recognition (NER) tasks, the IOB label sequence of each augmented text typically differs from that of the original, as the positions of named entities are likely to change during augmentation. Consequently, it is necessary to regenerate label sequences for the augmented texts. To achieve this, we employ a prompt-based approach: we provide the original text and its corresponding label sequence to LLMs, and then prompt LLMs to generate the appropriate IOB label sequence for the augmented text. The prompt template is shown below:

\begin{tcolorbox}[breakable, title=Prompt for IOB Label Generation]
\scriptsize
  \texttt{You are a professional named entity recognition (NER) annotation expert. Your task is to tokenize the given sentence, identify the named entities, and assign a corresponding BIO-format label and label ID to each token.} \\
  
  \texttt{This task includes only the following four types of entities: persons (PER), organizations (ORG), locations (LOC), and miscellaneous names (MISC).} \\
  
  \texttt{Use the following BIO labels and their corresponding label IDs: \{'O': 0, 'B-ORG': 1, 'B-MISC': 2, 'B-PER': 3, 'I-PER': 4, 'B-LOC': 5, 'I-ORG': 6, 'I-MISC': 7, 'I-LOC': 8\}.} \\

  \texttt{Please output the result strictly in the following format (only include these four lines):} \\
  \texttt{sentence: original sentence} \\
  \texttt{entities: ['token1', 'token2', ..., 'tokenN']} \\
  \texttt{labels: [BIO\_label1, BIO\_label2, ..., BIO\_labelN]} \\
  \texttt{IDs: [label\_id1, label\_id2, ..., label\_idN]} \\

  \texttt{Here is an example:} \\
  \texttt{sentence: \{example\_sentence\}} \\
  \texttt{entities: \{example\_entities\}} \\
  \texttt{labels: \{example\_labels\}} \\
  \texttt{IDs: \{example\_ids\}} \\

  \texttt{Now, please perform named entity recognition and annotation for the following sentence:} \\
  \texttt{\{input\_text\}} \\

  \texttt{Return only the result. Do not include any explanation or additional content.} \\
\end{tcolorbox}

\end{document}